\documentclass[letterpaper, 10 pt, conference]{ieeeconf}  % Comment this line out if you need a4paper

\IEEEoverridecommandlockouts                              % This command is only needed if
\usepackage{graphics} % for pdf, bitmapped graphics files
\usepackage{epsfig} % for postscript graphics files
\usepackage{mathptmx} % assumes new font selection scheme installed
\usepackage{times} % assumes new font selection scheme installed
\usepackage{amsmath} % assumes amsmath package installed
\usepackage{amssymb}  % assumes amsmath package installed
\usepackage{hyperref}
\usepackage{colortbl}
\usepackage{multirow}
\usepackage{multicol}
\usepackage{biblatex}
\usepackage{color}
\usepackage{pifont}
\usepackage[usenames,dvipsnames,table]{xcolor}
\hypersetup{
    colorlinks=true,
    linkcolor=MidnightBlue,
    filecolor=magenta,      
    urlcolor=MidnightBlue,
    citecolor=MidnightBlue,
} 
\usepackage[font=small]{caption}

\title{\LARGE \bf EyeSight Hand: Design of a Fully-Actuated Dexterous Robot Hand with Integrated Vision-Based Tactile Sensors and Compliant Actuation}

\author{
   \authorblockN{Branden Romero*, Hao-Shu Fang*, Pulkit Agrawal and Edward Adelson %
       \authorblockA{Massachusetts Institute of Technology \\
   {\tt\small brromero@mit.edu, fhs@mit.edu, pulkitag@mit.edu, adelson@csail.mit.edu}}
\thanks{* Equal contribution. Toyota Research Institute provided funds to support this work.}% 
}}
\usepackage{graphicx}
\usepackage{etoolbox}
\newcommand{\insertfig}{
\includegraphics[width=\linewidth]{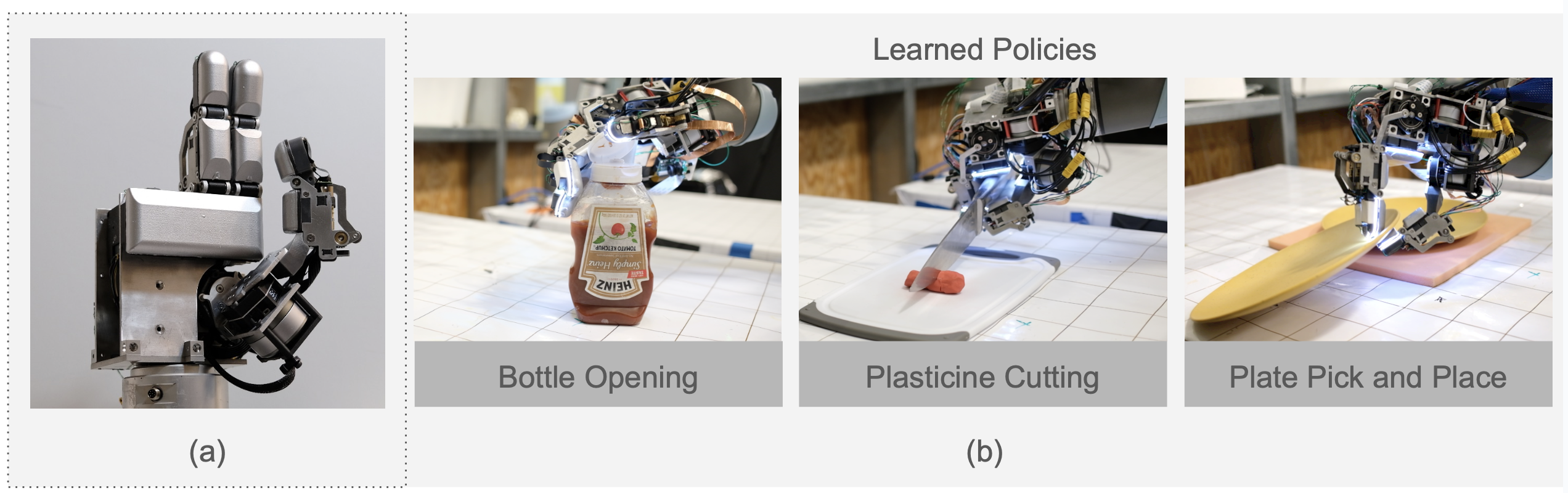}
\captionof{figure}{(a) Picture of the EyeSight Hand. (b) EyeSight hand is used to perfrom three tasks autonomously using imitation learning: bottle opening, plasticine cutting, and plate pick and place.}
           \label{fig:teaser}
}% define the image

\makeatletter
\apptocmd{\@maketitle}{\centering\insertfig}{}{}% insert the figure after authors
\makeatother

\begin{document}

\maketitle
\thispagestyle{empty}
\pagestyle{empty}
\addtocounter{figure}{-1}

\begin{abstract}
In this work, we introduce the EyeSight Hand, a novel 7 degrees of freedom (DoF) humanoid hand featuring integrated vision-based tactile sensors tailored for enhanced whole-hand manipulation. Additionally, we introduce an actuation scheme centered around quasi-direct drive actuation to achieve human-like strength and speed while ensuring robustness for large-scale data collection. We evaluate the EyeSight Hand on three challenging tasks: bottle opening, plasticine cutting, and plate pick and place, which require a blend of complex manipulation, tool use, and precise force application. Imitation learning models trained on these tasks, with a novel vision dropout strategy, showcase the benefits of tactile feedback in enhancing task success rates. Our results reveal that the integration of tactile sensing dramatically improves task performance, underscoring the critical role of tactile information in dexterous manipulation. 

\end{abstract}

\section{INTRODUCTION}
\begin{figure*}[]
  \centering
  \includegraphics[width=\textwidth]{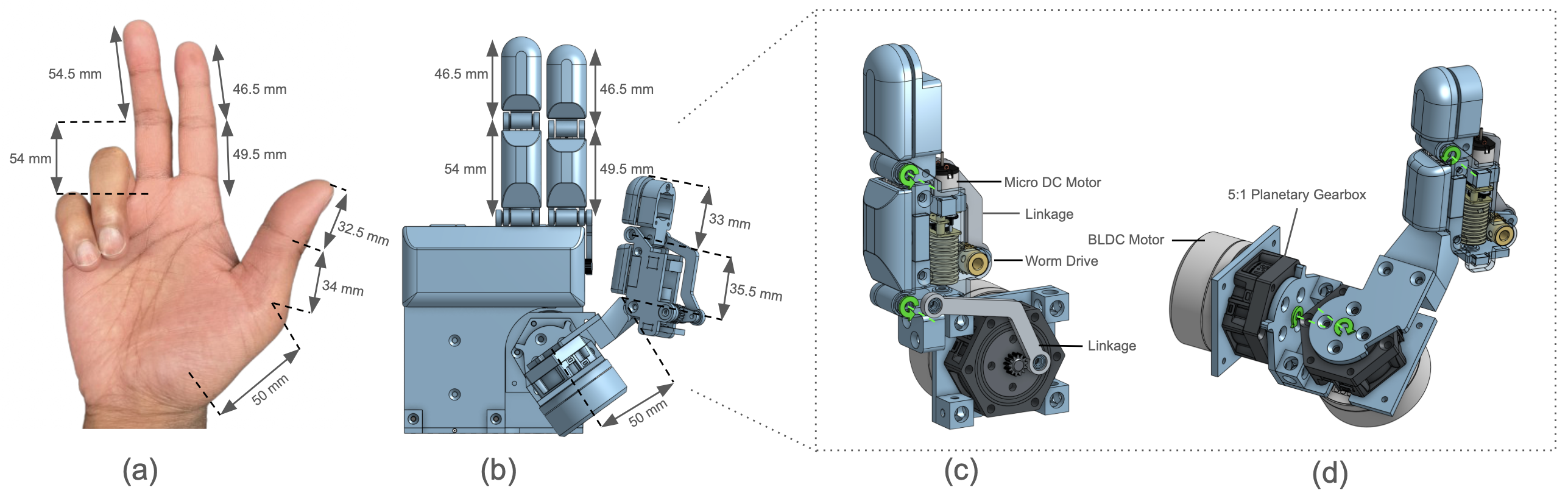}
  \caption{(a-b) Comparison of the dimension of a human hand with the dimensions of the EyeSight Hand. (c-d) Diagram illustrating the kinematics of the (c) index/middle finger and (d) thumb as well as the placements of the motors and transmissions. Axis of rotation is shown with green arrows. }
  \label{fig:mechanical}
\end{figure*}
The versatility of the human hand is a result of a unique blend of anatomical complexity that combines a complex kinematic structure, an array of muscles and tendons that provide compliance to manipulate objects with varying precision and delicacy, and a high density of tactile sensors in the skin to provide detailed feedback on object geometry and contact forces~\cite{jones}.

In various efforts to replicate the dexterity of the human hand with robotic hands, previous methods have captured only a subset of human hand functionality. For example, current fully actuated hands~\cite{leap} are usually not compliant, characterized by low backdrivability and/or high reflected inertia due to high gear reductions. Soft hands~\cite{rbo,zhou2019soft} that are compliant are often not fully actuated and difficult to control precisely. They also face challenges in incorporating dense tactile sensing. Existing hands equipped with dense tactile sensors~\cite{svelte,achu} typically have a limited number of degrees of freedom (DoF) and do not closely mimic human hand morphology. However, all these properties are crucial for robot learning: compliance makes the hardware robust to large-scale data collection, dense tactile sensing is important for precise manipulation, and a morphology similar to the human hand facilitates easy teleoperation and data collection.
%More over, with the rapid deployment data-driven methods, any implementation should be robust to large-scale data collection. 

In this work we introduce EyeSight Hand, a low-cost (less than \$2500), 7-DoF humanoid hand that is co-designed with actuation, tactile, and kinematics in mind. The key contributions of the EyeSight Hand are:
\begin{itemize}
  \item We exploit the benefits of quasi-direct drive actuation and how to integrate it into a humanoid robotic hand, to provide compliant actuation while having human level finger strength and speed. All while being robust enough for large-scale data collection.
  \item With vision-based tactile sensors emerging as a promising approach to robotic touch, we introduce a new vision-based tactile sensor GelSim(ple) that is easily simulatable and can be adapted to different morphologies while maintaining the sensitivity of previous GelSight sensors. We then integrate 8 of these sensors into the hand to enable sensitive whole-hand manipulation.
  \item Given the above two design choices, we manage to make the hand close enough to the morphology of human hand, in terms of kinematics and dimensions, which enables simple and intuitive teleoperation with minimal kinematic retargeting effort.
\end{itemize}
% by showing its wide applicability~\cite{fingerslam,swingbot,extrinsic,slip}With vision-based tactile sensors emerging as a promising approach to robotic touch by showing its wide applicability to perform contact localization and object reconstruction ~\cite{fingerslam}, estimating dynamical properties of an object through exploration ~\cite{swingbot}, estimating and controlling extrinsic contact ~\cite{extrinsic}, and detecting incipient slip~\cite{slip}, we introduce a new vision-based tactile sensor GelSim(ple) that is easily simulatable and can be adapted to different morphologies while maintaining the sensitivity of previous GelSight sensors. We then integrate 8 of these sensors into the hand to enable sensitive whole-hand manipulation.

To evaluate our hand, we propose three challenging tasks that that require complex non-prehensile manipulation, forceful tool-use, extrinsic force sensing, and precise dexterous manipulation, including bottle opening, plasticine cutting, and plate pick and place. Our hand can be teleoperated to accomplish these tasks without much effort, and works robustly during hundreds of trials. Finally, we train imitation learning models on these tasks, where a vision dropout strategy is proposed for the model to better utilize the tactile sensing, and quantitatively show the benefits of our dense tactile sensing.
% we validate the hand through these task
% no matter IL or RL, hand would be easy to collide with env, thus would need backdrivability
% move fast, improve teleop efficiency
% similar to human hand, good for teleop
% good tactile would make it easier to learn

% bom less than \$2500

\section{BACKGROUND AND RELATED WORK}
\subsection{Humanoid Robotic Hands}
There is a plethora of research trying to replicate the versatility of human hands to robotic hands. One of the most notable attempts is the commercially available Shadow Hand, a highly dexterous but costly platform. Whose tendon-driven approach reduces the reliability of platform significantly. Hands whose motors directly actuate the joint have been developed like the commercially available Allegro Hand as well as research efforts like ~\cite{leap,manus}. However, they rely on small highly geared servos, that have a tendency to overheat under continuous load, and the high reflected inertia of the gearbox leads to high resistance to external forces which increases risk of breakage. Recently,~\cite{ilda} has shown potential of linkage-based designs particularly in terms of robustness to large forces, however the actuation is slow making it unsuitable for dynamic manipulation and potentially difficult to teleoperate. Although not specifically a humanoid hand, ~\cite{trifinger} illustrates the potential of quasi-direct drive actuators for robust and dynamic manipulation. We explore such potential for humanoid hand in this paper.

\subsection{Hands with Vision Based Tactile Sensors}
While hands like~\cite{ilda,manus} have tactile sensing, they are low resolution compared to vision-based tactile sensors. To enable dense tactile sensing, researchers have proposed incorporating vision-based tactile sensors, such as GelSight~\cite{yuan2017gelsight}, into robotic hands. Due to hardware constraints, early attempts~\cite{wedge, donlon2018gelslim, allegro, padmanabha2020omnitact} were typically flat or round and only mounted on the fingertips of hands. Recently, some works~\cite{she2020exoskeleton, wilson2020design, liu2022gelsight, zhao2023gelsight} have explored integrating the GelSight sensor across the entire finger. Existing research has focused on parallel grippers with only 2 or 3 DoF. In this work, we propose a novel design of the GelSight sensor and integrate it with a self-designed 7 DoF hand.

\subsection{Learning for Dexterous Hand}
Learning manipulation with a dexterous hand is challenging due to the complexity of managing its many degrees of freedom and the need for precise sensing under severe occlusion. Early algorithms mainly conducted reinforcement learning with ground-truth~\cite{zhu2019dexterous,rajeswaran2017learning,kumar2016optimal,nagabandi2020deep} or estimated state information~\cite{andrychowicz2020learning,akkaya2019solving}. To deal with high-dimensional observation, teacher-student learning~\cite{chen2022system} or representation learning~\cite{huang2021generalization} are proposed, showcasing results
of transferring to real-world scenarios in hand rotation~\cite{chen2023visual}. Beyond visual observation, some recent work~\cite{qi2023general, yin2023rotating, yuan2023robot, suresh2023neural} also explored incorporating tactile observation for in-hand rotation tasks. These methods are trained with reinforcement learning in simulation due to the demand for large-scale data. Few work have explored direct imitation learning in the real world since it is unclear how much data is needed to train a good policy, and the hardware issues like non-compliant also make it difficult to teleoperate and collect data. Recent attempts in this direction need additional data source including internet-scale video and augmented simulation demo to pretrain the policy~\cite{shaw2023leap, wang2024cyberdemo}, or play data to learn a representation model~\cite{guzey2023dexterity}. In this work, we demonstrate that by incorporating rich tactile perception, we can achieve efficient imitation learning for dexterous manipulation on several challenging contact-rich tasks in the real world.

% previous work is mainly RL, because it's hard to teleop or collect data, due to hardware constraints
% works in IL, few, because of hard, due to lack of tactile, vision only occluded heavily
\section{Mechanical Design}
\label{sec:design}
Given the recent achievements in machine learning, the goal is to design a hand that is robust enough for large scale data collection, intuitive enough to perform teleoperation for imitation learning, and a platform that can potentially take advantage of data of humans performing task. In this section we explore the design decisions to achieve that goal.

\subsection{Kinematics}
\label{sec:design}
\begin{figure}[t]
  \centering
  \includegraphics[width=0.33\textwidth]{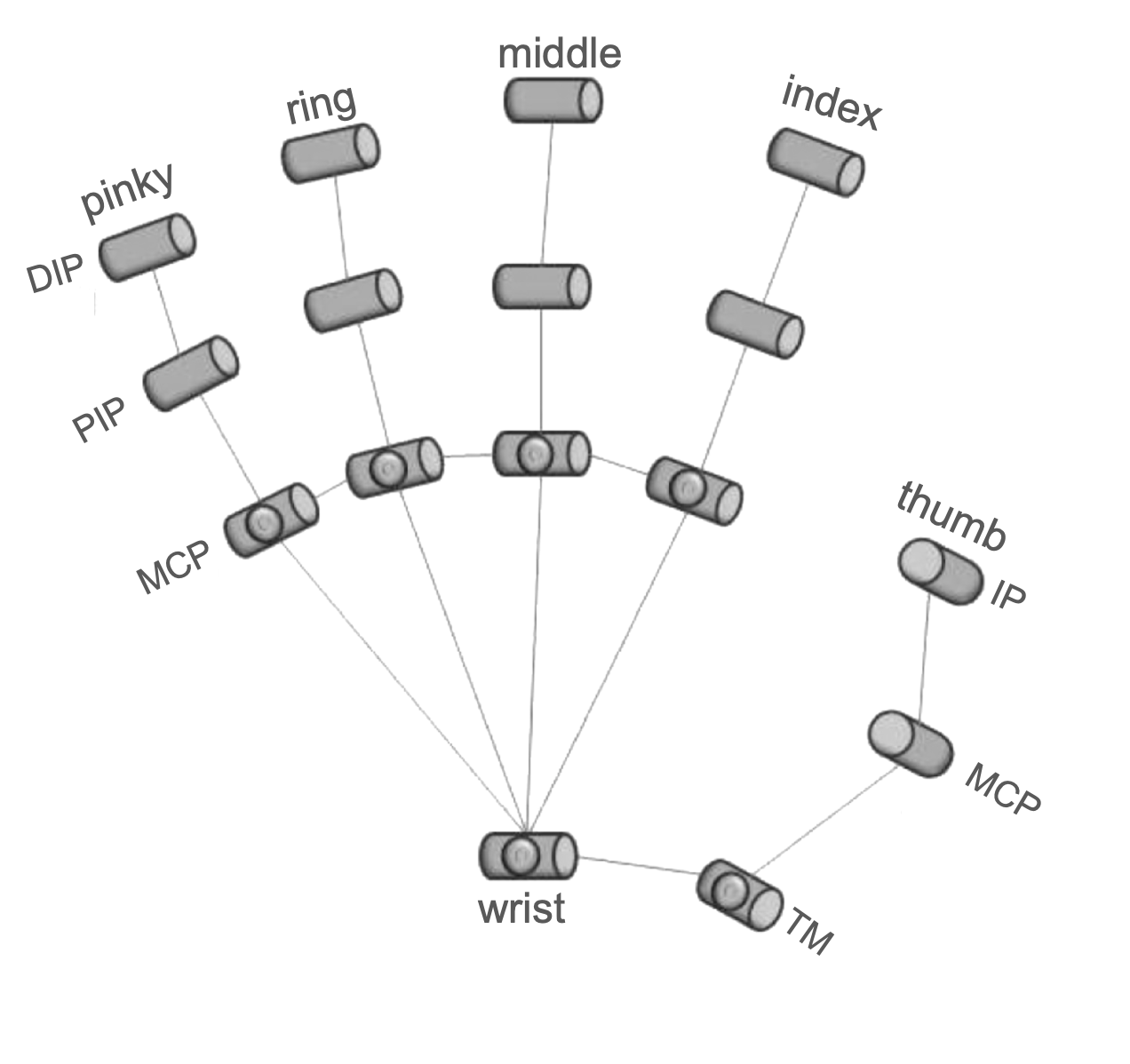}
  \caption{Simplified kinematics of the human hand~\cite{human_kin}.}
  \label{fig:humkin}
  \vspace{-0.1in}
\end{figure}
Naturally, to achieve intuitive teleoperation and to best take advantage of human data, we would like to close the embodiment gap and try to closely mimic the kinematics of human hand, as illustrated in Fig.~\ref{fig:mechanical}, as much as possible. However, in the design of the EyeSight Hand we make several compromises to fit in other features. Mainly,  the hand differs from human kinematics (as shown in Fig.~\ref{fig:humkin}) in the metacarpophalangeal (MCP) joints and distal interphalangeal (DIP) joints of the index and middle finger, and the MCP joint of the thumb. Where, the DIP joints and the lateral axis of the MCP of the index and middle finger are missing, and the MCP joint of the thumb is missing. 

\subsection{Actuation}
The actuation scheme is motivated by how compliance of the human hand. In particular, the observation that given an external force to the fingertip, the DIP and proximal interphalangeal (PIP) joints of the finger remain relatively static, while the flexor axis of the MCP joint back drives. 
Therefore, the hand takes use of two types of actuators, a 30:1 N20 Micro DC Motor driving a 20:1 worm gear reduction, and a TMotor GL40 KV210 brushless DC (BLDC) motor with a 5.23:1 planetary reduction. These actuators are placed in series for design simplicity, where for the middle and index finger the worm drive actuates the proximal interphalangeal PIP joint, while the BLDC motor actuates the flexor axis of the MCP joint. Meanwhile for the thumb the worm drive actuates the interphalangeal (IP) joint, while the BLDC motor actuates both axis of the trapeziometacarpal (TM) joint. These actuator either actuate the joints by directly driving them or via a four bar linkage, making our hand extremely robust compared to tendon driven ones.

The selection of these specific actuators was based on their durability. We choose the worm gear, due to its high reduction in a small factor, and its self-locking/ non-backdriving capability. The high reduction is particularly advantageous as it allows us to use a relatively large tooth size robust to high forces. Then due to self-locking, all of the forces get transmitted to the BLDC motor. Unlike the multistage spur gear servos used in many of dexterous hands like~\cite{leap}, the low reduction of the BLDC motor means less friction in the transmission and lower reflected inertia. This leads to less resistance on the actuator to external forces leading to decreased risk of breakage. Also, the planetary gear evenly distributes the load across multiple gears, leading to increased lifespan and reliability. This load distribution also contributes to the ability to handle high shock loads.

 This actuation scheme allows for extremely nimble and strong fingers, with an estimated 19N continuous fingertip force, max fingertip force 57N, and a maximum speed of 420 RPM. Beyond exerting high forces, high torque actuators have the benefit of being less likely to overheat from sustained use, as the continuous operating torque of these actuators greatly exceed the stall torque of the motors used in~\cite{leap} and the commercially available Allegro Hand.

\subsection{Electronics}
The BLDC actuators are driven by a remotely place MJBOTS Moteus N1 Brushless Controller, and are equipped with AMS AS5047P 14-bit magnetic encoder to provide rotor position. Communication with the BLDC actuators happen over 5MBPS CAN-FD at 1kHz. The BLDC controller is equipped with current sensing, allowing for torque estimation from current ~\cite{torquecur}. The three micro DC motors are driven by TI DRV8876 controlled by a single Teensy 4.1 and communicate over serial. 

\begin{figure}[t]
  \centering
  \vspace{0.1cm}
  \includegraphics[width=0.5\textwidth]{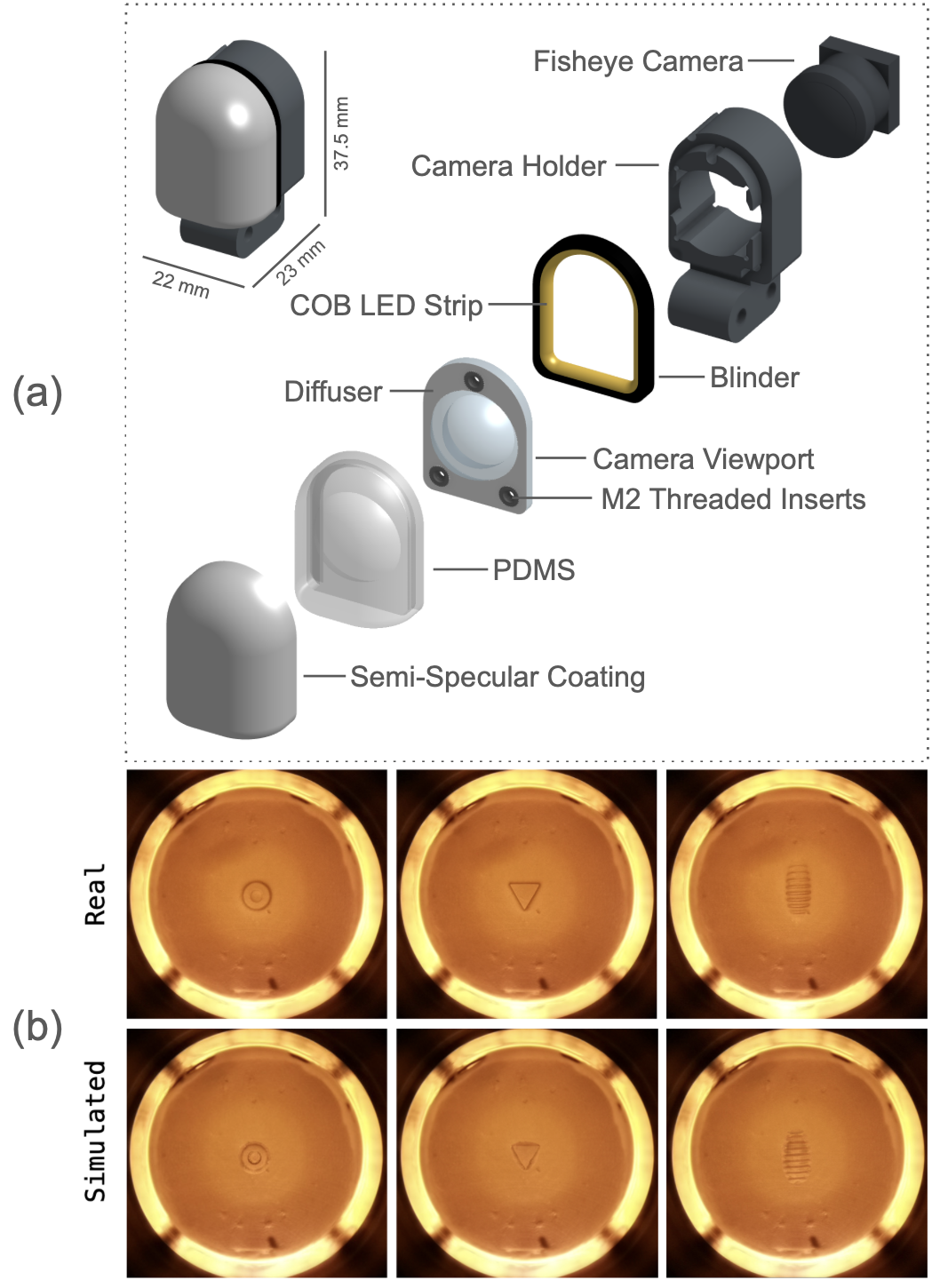}
  \caption{(a) Dimension and exploded view of the GelSim(ple) sensor at the tip of the thumb. (b) Example raw outputs of the GelSim(ple) and the approximate simulation of the corresponding deformation.}
  \vspace{-0.1in}
  \label{fig:tactile}
\end{figure}

\section{TACTILE SENSOR DESIGN}
One of the main challenges in designing the EyeSight Hand was in fully integrating vision-based tactile sensing into the hand. The EyeSight Hand consists of seven tactile surfaces, each with varying morphologies, streaming a total of eight images, one for each finger segment, and two for the palm. So the question becomes, how do we come up with a design framework adaptable to different sensor morphologies, that maximizes data throughput, and also enables us to simulate the sensor for sim2real approaches. Therefore, in this work we introduce GelSim(ple) a novel variant of the GelSight Sensor. 

\subsection{Illumination}
The initial goal of GelSight was to perform 3D reconstruction of the sensing surface via photometric stereo. This was achieved by illuminating the sensing surface with 3 different colored directional light sources spaced as evenly apart as possible around the sensor while minimizing their angle of elevation to the sensing surface~\cite{yuan2017gelsight}. While this was easy to implement for flat sensors, adapting this strategy to other geometries is difficult and results in various compromises~\cite{360finger,svelte,allegro}. However, findings from \cite{wedge} suggested that a learning-based approaches can infer the 3D geometry of the sensing surface with single colored directional illumination. Furthermore, when it came to simulating GelSight Sensors, directional lighting meant that not only did you have to find a mapping from surface normal to RGB, you also had to accurately simulate shadows~\cite{taxim}, which is costly. Given these findings, GelSim(ple) uses a new illumination strategy, which side illuminates the perimeter of a diffuser to provided non-directional overhead lighting to a semi-specular sensing surface as shown in Fig.~\ref{fig:tactile}(a), which still clearly gives geometric information about the object in contact, but also eliminates casted shadows (Fig.~\ref{fig:tactile}(b)). Furthermore, this allows us to increase data throughput by only transmitting a single channel image. To simulate the sensors, pixel-wise inference is performed via a multilayer perceptron (MLP) that takes as input surface normal and viewing direction augmented with positional encoding of a simulated deformation, and outputs the estimated RGB value~\cite{difftactile}.

\subsection{Implementation Details}
GelSim(ple) (Fig.~\ref{fig:tactile}(a)) uses an Arducam B0286 IMX219 Camera Module with a 220 degree fisheye lens. The camera is press fit into a 3D printed finger segment body, or palm body, that fastens to the sensing module via M2 screws. The sensing module consist of a cut to length SuperLightingLed Narrowest 3mm COB LED White Light Strip formed into the desired geometry and adhered to a 3D printed blinder. The lighting module is then pressed onto a clear rigid 2mm acrylic backing with a spherical view port for the fisheye camera to observe the sensing surface with minimal optical distortion. The acrylic backing has a 3M 3635-70 diffuser on the surface opposite of the camera, and has several M2 threaded inserts to attach it to the camera holder. SILICONES, INC. XP-565 is molded onto the rigid backing and coated with Print-On® Clear Silicone Ink mixed with a 13 micron non-leafing silver dollar aluminum pigment. Each camera is connected remotely to its camera driver board via a Arducam B0439 Sensor Extension Cable. The hand has two Arducam B0396 CamArray Hats allowing 8 synchronized camera modules to be interfaced to two Raspberry Pi 4. A single channel image is transmitted from each Raspberry Pi via ZeroMQ at 60Hz at 640x480 to the host computer.
\section{EXPERIMENT SETUP}
\label{sec:exp}

\begin{figure*}[ht]
  \centering
  \vspace{0.1cm}
  \includegraphics[width=\textwidth]{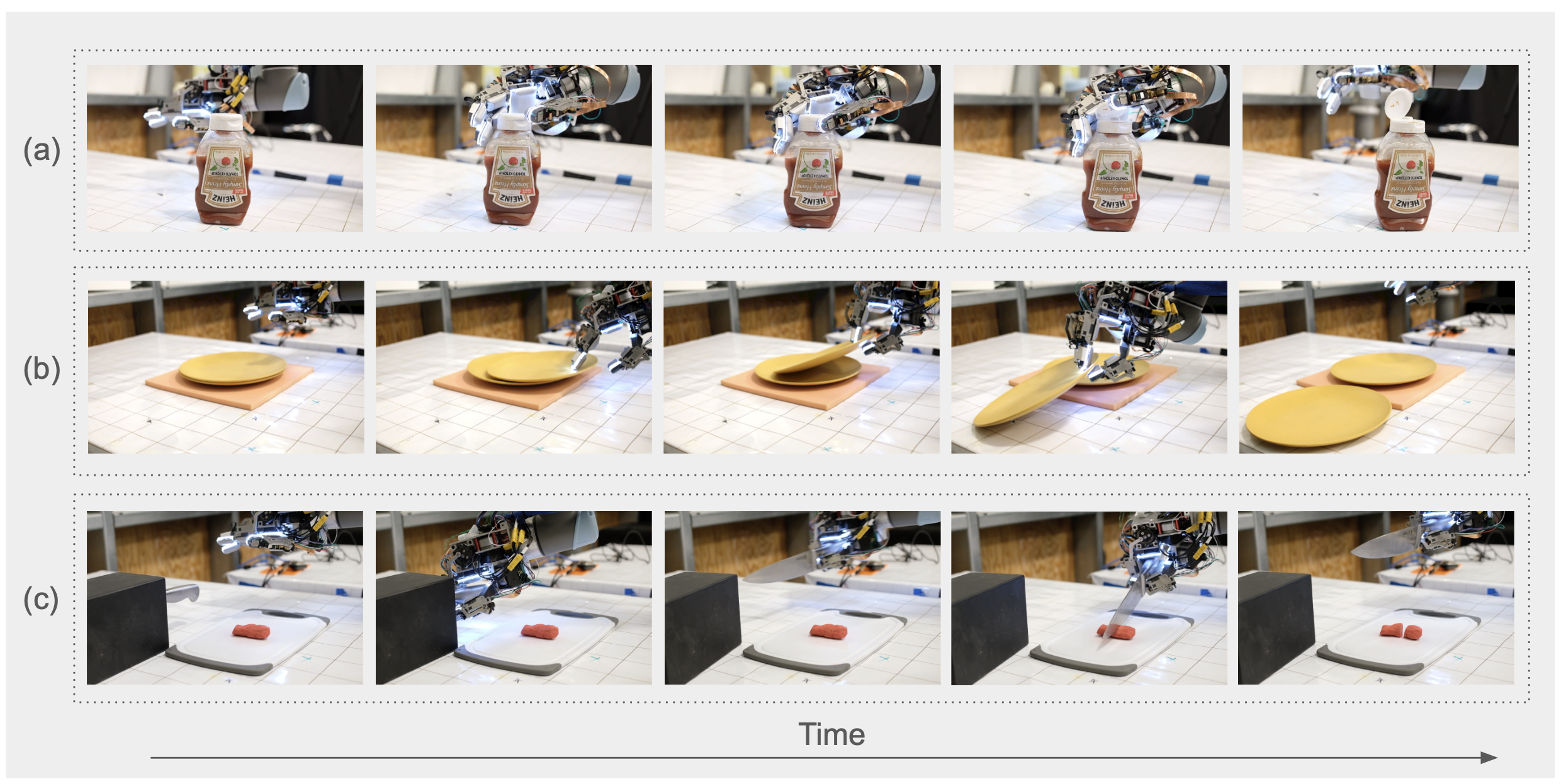}
  \caption{Illustrations of the evaluation task. (a) The robot performs lid opening, where the robot must first approach the bottle, then constrain it with the index, middle finger, and palm, and then use the thumb to open the bottle. (b) The robot must approach a set of stack places, and then slide one over the other until it is able to grasp the plate. It then picks the plate and places next to the other plate. (c) The robot must remove a knife from a box, and use it to cut a plasticine roll fully.}
  \label{fig:task}
\end{figure*}

To validate the performance of the EyeSight Hand, we perform imitation learning on three tasks: cutting plasticine, picking and placing stacked plates, and opening bottles. We hope this will demonstrate the hand's capabilities and the importance of tactile sensing in performing a diverse set of tasks, which require complex non-prehensile manipulation, forceful tool use, extrinsic force sensing, and precise dexterous manipulation. In this section, we describe the system used to collect data, provide a breakdown of each task, and discuss our imitation learning approach.

\subsection{Teleoperation}
\label{sec:teleoperation}
Teleoperation is performed using the 3D Guidance trakStar tracking
system. The operator wears a glove equipped with four
magnetic sensor probes, three fastened near the finger nails of the
thumb, index, and middle, while the other placed on the
back of the palm. Each of
these sensors provide a 6-DoF pose. To estimate the commanded joint angles, we perform kinematic retargeting
via Task Space Vector (TSV) optimization~\cite{dexmv,dexpilot,tsv}. A
human hand model is fully determined by the four sensor poses, e.g.,
$\{\mathit{H}_1, \mathit{H}_2, \dots, \mathit{H}_4\}$. Accordingly,
the URDF of the robotic hand specifies a set frames that are
attached to the finger tips and the palm, e.g.,
$\{\mathit{R}_1, \mathit{R}_2, \dots, \mathit{R}_4\}$ which is
parameterized by the joint angles of the robot $Q(t)$ at time $t$. For
each set of coordinate frame we create a set of task space vectors, where
each task space vector points from the origin of one coordinate system
to another denoted as $r_{ij}(Q(t))$ and $h_{ij}$ for the robot hand
and human hand respectively in which $i=1,2,\dots,4$,
$j=1,2,\dots,4$, and $i\neq j$. Then perform the following optimization routine:
\begin{equation}
  \label{eq:optim}
    \min_{Q(t)} \sum_{i\neq j} \left\lVert r_{ij}(Q(t)) - h_{ij} \right\rVert ^2 + \alpha \left\lVert Q(t) - Q(t^{\prime}) \right\rVert ^2,
\end{equation}
in which $t^{\prime}$ is the previous time step and $\alpha$ is a
weight to address the significance of the change of the robotic hand
pose. In terms of implementation, Eq.~\ref{eq:optim} \ is solved
using Sequential Least-Squares Quadratic Programming
(SLSQP)~\cite{slsqp,nlopt}. The optimization routine was written in
JAX~\cite{jax} and makes use of the Automatic Differentiation for
rigid-body-dynamics AlgorithMs library~\cite{adam}, to feed the
forward and backwards solution of the objective to SLSQP in a timely
manner. Finally, the result of the retargeting is used to directly
command the joint angles of the robot hand. The teleoperation system
operates at 125hz.

\subsection{Task Specification}
\label{sec:taskspec}
For each task we collect proprioceptive data from the EyeSight Hand and the UR5 robot, images from a global camera,  images from a fisheye camera placed at the wrist of the hand, and images from the tactile sensors at 30Hz. For each task we collect 100 demonstrations. 

\subsubsection{Bottle Opening}
For this task (Fig.~\ref{fig:task}(a)) a ketchup bottle is randomly placed within an area with slight perturbations to its initial orientation. The bottle is not fixed to the table so the robot must first approach the bottle and apply an appropriate constraint on the bottle so that it will not shift during manipulation, by applying downward force on the bottle's body. The thumb tip then comes in contact with the lip of the bottle opening and forces it open. The robot must then fully swing the lid open. If the lid is not fully open it does not count as a success. 

\begin{figure}[t]
  \centering
  \includegraphics[width=\linewidth]{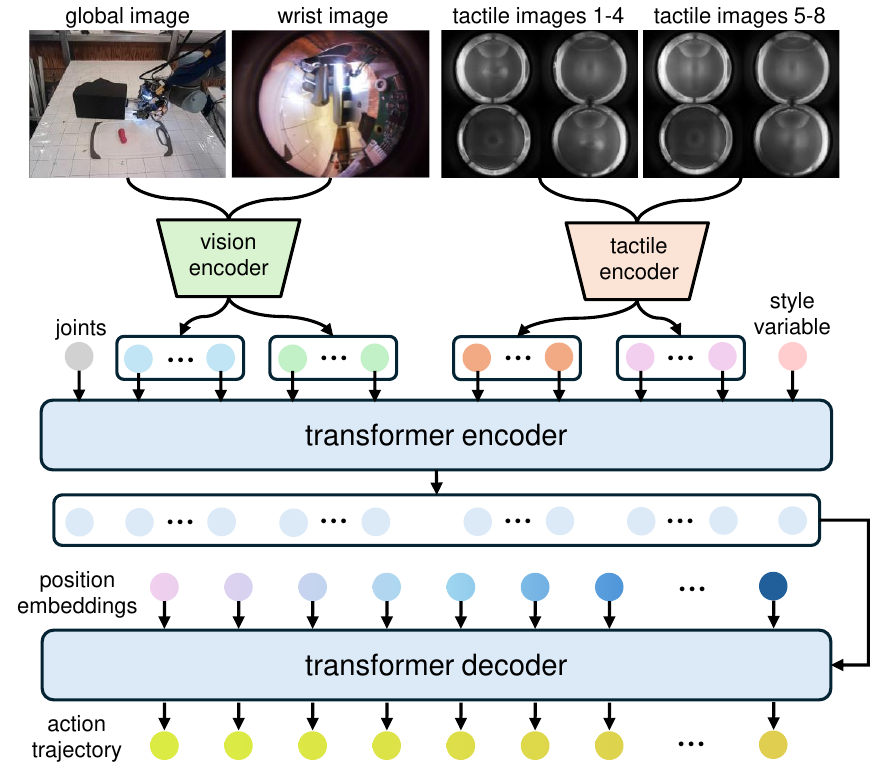}
  \caption{Illustration of our input to the policy along with the architecture of the policy network.}
  \label{fig:act-network}
\end{figure}

\subsubsection{Plate Pick and Place}
For this task (Fig.~\ref{fig:task}(b)) two stacked plates are placed on top of a hard foam block for slight friction. At the beginning of each episode, the robot must approach the stacked plates and make contact with the top plate. The robot then gently slides the top plate over the bottom plate until it is in a graspable position. Too much force applied to the top plate will result in both plates sliding. The top plate is then lifted and gently placed next to the other plate. Initial positions of the stacked plates and foam block are slightly perturbed.

\begin{figure*}[t]
  \centering
  \vspace{0.1cm}
  \includegraphics[width=\textwidth]{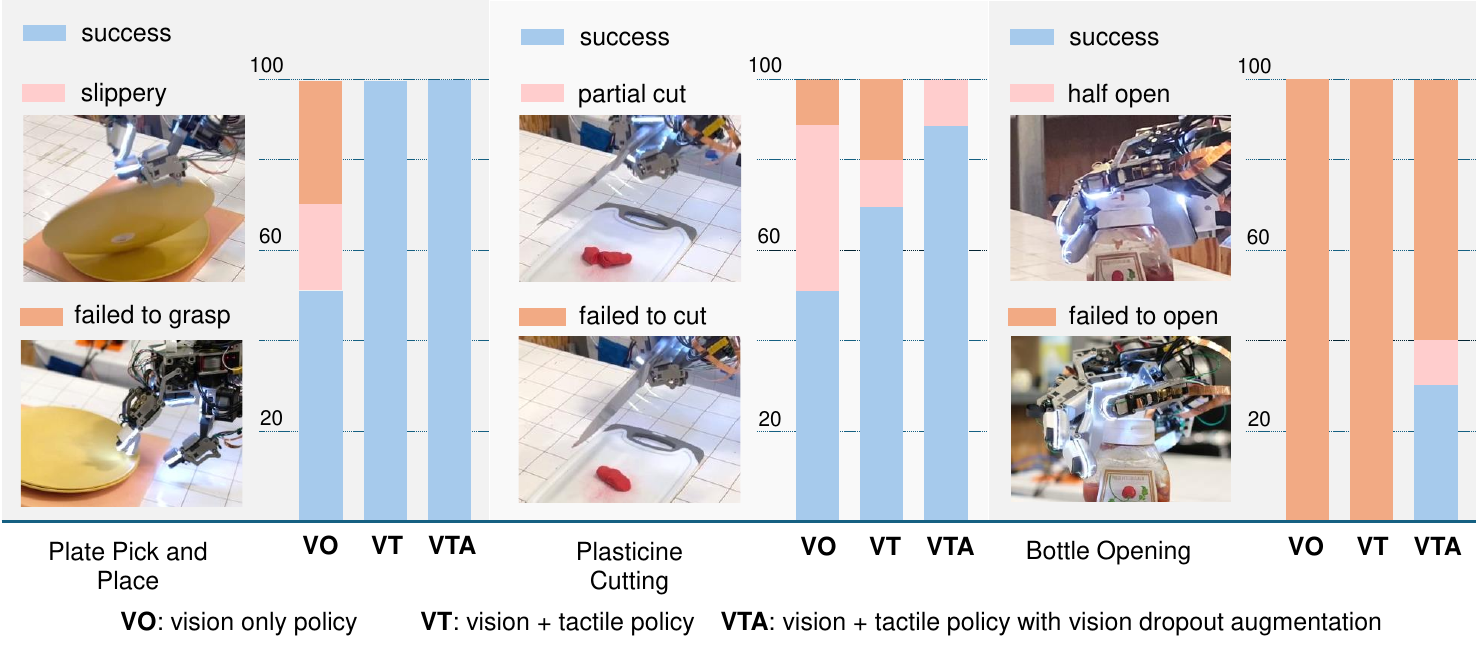}
  \caption{Success rates of the various variants of our trained policies across different tasks, along with a detailed breakdown of the failure cases.}
  \label{fig:learning_results}
\end{figure*}

\subsubsection{Plasticine Cutting}
For this task (Fig.~\ref{fig:task}(c)) The robot must approach a knife stored in a box and pull it out of the box. The robot then goes and cuts a plasticine roll. The plasticine roll must be cut fully in half for the task to be considered successful. The knife box and cutting board are not fastened to the table, so excessive forces will cause them to slide. The initial position of the knife box is perturbed as well as the shape and position of the plasticine roll.

\subsection{Imitation Learning}
To learn dexterous manipulation, we adopt Action Chunking with Transformers (ACT)~\cite{zhao2023learning}. ACT is a generative model structured as a conditional variational autoencoder (CVAE), where the CVAE encoder learns a hidden style variable mapped to a diagonal Gaussian, and the CVAE decoder learns to predict the future action trajectory given the current observations and the style variable. This design accommodates the multi-modal nature of human-teleoperated action trajectories. For more details, we refer readers to ~\cite{zhao2023learning}. In Fig.~\ref{fig:act-network}, we illustrate our input format and the architecture of the CVAE decoder, which serves as the policy network. The images from both the global camera and wrist camera are processed through a shared vision encoder to obtain their visual features. For the eight tactile images, we divide them into two subsets and concatenate every four images into a $2\times2$ super-image. The resulting two super-images are then processed through a shared tactile encoder. Both visual and tactile features, concatenated with the robot joint positions and the style variable, are fed through a transformer network to predict the future action trajectory in joint space.

We train three variants of ACT. The first, referred to as vision-only ACT, takes only visual images from the global and wrist cameras as input, without employing the tactile encoder branch during both training and testing phases. The second, referred to as vision-tactile ACT, incorporates both visual and tactile images. However, during our experiments, we observed that the vision-tactile ACT sometimes ignored the tactile signal. Therefore, we propose a third version of ACT. In this version, during training, we introduce a random dropout for the input to the vision encoder—setting the entire images from the global and wrist cameras to zeros with a 30\% chance. During testing, the input images are processed without dropout.

For our implementation, we resize all visual images and tactile super-images to $320 \times 480$ and adopt ResNet-18~\cite{resnet} for both the vision and tactile encoders. The joint positions vector is 13-dimensional, consisting of 6 degrees of freedom (DoF) for the UR5 arm and 7 DoF for the hand. We predict the action trajectory for the next 20 timesteps, resulting in a $20\times13$ tensor. The prediction frequency is set to 15Hz. We train our policy on an Nvidia A6000 GPU with a batch size of 48. The learning rate is set to $\texttt{1e-5}$, and all policies are trained for 500 epochs. Following~\cite{fang2023low}, we set the KL weight, hidden dimension, and feedforward dimension of ACT to 10, 1024, and 6400, respectively. During testing, our policy operates at 15Hz.

\section{EXPERIMENTAL RESULTS}
We compared the results of policies trained with different strategies, rolling them out on the real robot across the three tasks introduced in~\ref{sec:taskspec}. Each experiment was conducted over 10 trials. In Fig.~\ref{fig:learning_results}, we show the success rates of different policies on various tasks and a breakdown of the failure cases.

For the plate pick and place task, the vision-only policy achieved a mere 50\% success rate. Incorporating tactile sensing increased the success rate to 100\%. Without tactile sensing, there were two main failure modes: The first occurred when the robot attempted to slide the first plate but either failed to make contact or applied insufficient force, leading to repeated unsuccessful sliding attempts without moving the plate. The second failure mode involved the robot grasping the plate without applying enough force, resulting in an unstable grasp and causing the plate to slip from its hand during movement.

For the plasticine cutting task, the vision-only policy also achieved a mere 50\% success rate. Adding tactile sensing in a conventional manner improved the result to 70\%. After adopting our vision dropout training strategy, the policy could achieve a success rate of 90\%, perfectly grasping the knife and cutting the plasticine in half. Analyzing the failure cases, we found that without tactile sensing, the policy struggled to fully cut through the plasticine, as the visual difference between a partial and full cut is subtle in images. Meanwhile, in successful cases, the robot could apply excessive force, causing the cutting board to move with the knife. Incorporating tactile sensing significantly reduced the occurrences of partial cuts and prevented excessive force application during cutting, thus not moving the cutting board. The vision dropout training strategy further enhanced the policy's reliance on tactile sensing, improving the robustness of its execution.

The bottle opening task was particularly challenging. Both the vision-only policy and the vanilla vision-tactile policy failed in all trials to open the bottle, either attempting to open the lid without making contact or applying force in the wrong direction, causing the thumb fingertip to slip away from the lid. When adopting the vision dropout training strategy, the policy achieved a 30\% success rate. The reason is that vision provides limited cues for the lid opening process, and the dropout strategy encourages the network to rely more on tactile sensing when predicting actions.

% all policies can grasp, but not all can cut well. show the importance in contact-rich manipulation
\section{CONCLUSION}
\label{sec:conclusion}
In this work, we present the Eyesight Hand, a fully-actuated 7-DoF hand equipped with high-resolution tactile sensing and compliant actuation. Our design renders the hand suitable for the robot learning community due to its rich sensing capabilities, robustness to collisions, and morphology closely resembling that of a human hand. We develop an imitation learning baseline and demonstrate that the rich tactile sensing from our hand significantly enhances performance on several challenging real-world tasks. In the future, we aim to extend our hand with additional fingers and explore algorithms to better utilize the rich tactile sensing.

\section{ACKNOWLEDGEMENT}
The research was sponsored by the Army Research Office and was accomplished under ARO MURI Grant Number W911NF-23-1-0277. The views and conclusions contained in this document are those of the authors and should not be interpreted as representing the official policies, either expressed or implied, of the Army Research Office or the U.S. Government.
\printbibliography
% \bibliographystyle{IEEEtran}
% \bibliography{bibliography}

\end{document}